\documentclass[preprint,12pt]{elsarticle}

\usepackage{amssymb}
\usepackage[numbers]{natbib}
\usepackage{graphicx}
\usepackage{amsmath}
\usepackage{amsthm}
\usepackage{url}
\usepackage{etoolbox}
\usepackage{float}
\usepackage{hyperref}
\usepackage{amsthm}
\usepackage{parskip}
\usepackage{lineno}

\journal{Information Fusion}

\begin{document}
\renewcommand*{\today}{November 28, 2022} 
\begin{frontmatter}



\title{Information Fusion via Symbolic Regression: A Tutorial in the Context of Human Health}


\author[l1]{Jennifer J. Schnur} 
\author[l1]{Nitesh V. Chawla \corref{cor1}} 

\address[l1]{Lucy Family Institute for Data and Society, Department of Computer Science and Engineering, University of Notre Dame, 46556, IN., USA}

\cortext[cor1]{Corresponding Author: nchawla@nd.edu}

\begin{abstract}
This tutorial paper provides a general overview of symbolic regression (SR) with specific focus on standards of interpretability.  We posit that interpretable modeling, although its definition is still disputed in the literature, is a practical way to support the evaluation of successful information fusion. In order to convey the benefits of SR as a modeling technique, we demonstrate an application within the field of health and nutrition using publicly available National Health and Nutrition Examination Survey (NHANES) data from the Centers for Disease Control and Prevention (CDC), fusing together anthropometric markers into a simple mathematical expression to estimate body fat percentage. We discuss the advantages and challenges associated with SR modeling and provide qualitative and quantitative analyses of the learned models.
\end{abstract}

\begin{keyword}
symbolic regression \sep interpretable modeling \sep information fusion \sep machine learning \sep mathematical representation 

\end{keyword}

\hypersetup{pdfauthor={Nitesh V. Chawla}}
\end{frontmatter}


\section{Introduction}
\label{intro}
Building predictive models using empirical data has become an essential part of analysis in nearly every field. As machine learning (ML) research has expanded, modeling approaches have reached beyond traditional regression-based methods toward approaches that can capture more complex, non-linear relationships hidden within data. One key example is deep learning (DL), which has demonstrated remarkable performance in a variety of contexts \cite{dong2021survey}. While rudimentary approaches, such as linear regression, often suffer from oversimplification of the relationships between features and target variables, DL approaches can make up for this at the expense of intrinsic interpretability, known as the ``black box'' effect, in which specific feature contributions to the model's predictions remain obfuscated. Within high-risk fields, such as medicine, finance, or criminal justice, model interpretability is especially crucial; the user must be able to ascertain why a certain decision should be chosen over another in order to maintain transparency and accountability \cite{felzmann2020towards}. Additionally, broader scientific modeling research may also benefit from a more flexible and interpretable approach, as the goal of such research is not only to make accurate predictions about various phenomena but also to understand the rules or patterns that intelligent systems are designed to uncover. Only by analyzing relationships between features and target variables can new knowledge be discovered and, importantly, used to make decisions that benefit society long-term. 

Symbolic Regression (SR) refers to the task of learning an optimal mathematical expression that best estimates a continuous target variable. While most regression approaches rely on tuning weights within a predefined model structure, a key component of the SR problem involves searching for the structure of the model itself by fusing together features, mathematical operators, and constants into a single function. The form of the expression has been learned through a variety of methods (see Section \ref{sr_overview}), ranging from biology-inspired genetic improvements to deep neural network policy representations. Various SR methods have been bench-marked \cite{la2021contemporary, egklitz2020BenchmarkingSS, Orzechowski2018, vzegklitz2021benchmarking} on problems that possess real ground-truth solutions, in addition to problems without known analytical forms. The problem has even extended to unsupervised learning through implicit function representations \cite{Schmidt2010SymbolicRO}, and has been shown to work well on small datasets \cite{wilstrup2021symbolic}. The key advantage of SR, in addition to non-linear pattern discovery and a high degree of portability, is the intrinsic interpretability that accompanies explicit mathematical model representation, allowing users to discern successful information fusion, especially within high-stakes applications.

Some of the most notable applications of SR have been performed in the context of physical systems, in which natural laws are derived from observational data \cite{schmidt2009distilling}, resulting in state-of-the-art software such as Eureqa \cite{dubvcakova2011eureqa} and AI Feynman \cite{udrescu2020ai, udrescu2006ai}. While SR has been heavily explored within the natural sciences \cite{vaddireddy2020feature, Wang2019SymbolicRI, murari2014symbolic, kabliman2021application, neumann2020new}, it has been understudied as means for fusing data from multiple contexts, despite its potential. For instance, in the healthcare domain, a patient's outcome regarding a certain disease or condition may not only rely on genetic components, patient behaviors, or clinical markers; it may also depend on demographic and socioeconomic variables that may impact the patient's environment, and hence his or her risk for developing said outcome. A few SR-based health studies have shown that disease risks \cite{goyal2022symbolic, christensen2022identifying} and treatment optimizations \cite{Virgolin2018} are modified by factors that can be fused into a single, concise expression. Through future application of SR, we may connect the multi-contextual features that encompass such risk, thereby uncovering holistic laws within the domain of human health. These SR models will not only allow researchers to explicitly understand associations between patient features and outcomes of interest, but also interactions between the features themselves, which lies at the core of information fusion principles. 

In this work, we provide an overview of SR methods and illustrate the means by which the resulting models meet various interpretability standards. We also demonstrate an application of SR in the context of human health, using as a case study the National Health and Nutrition Examination Survey (NHANES) database \cite{Nhanes}, collected by the Centers for Disease Control and Prevention (CDC) from 2017-2018. In this demonstration, we use off-the-shelf SR software to find functions that predict total body fat percentage in adults by taking into account biomarkers that have been used as measures of nutritional status, (see Section 3 for methodology details). Total body fat percentage, obtained via dual-energy x-ray absorptiometry (DXA), is widely used to measure body composition \cite{heymsfield1989dual}; however DXA is not always available in resource-constrained settings. Typically, in the absence of DXA, body mass index (BMI) \cite{keys1972indices}, defined as 

{\centering
\begin{equation}
    BMI = \frac{w}{h^2},
    \label{eqn:oldBMIeqn}
\end{equation}
}

\noindent where $w$ represents weight in kilograms (kg) and $h$ represents height in meters (m), is used as a screening tool, since it is moderately correlated with body fat; although it does not measure body fat percentage directly \cite{garrow1985quetelet, CDC_BMI}. While BMI has been demonstrated as a decent predictor of all-cause mortality and various adverse health outcomes \cite{flegal2013association}, there are many known drawbacks regarding the use of BMI as an indicator. Specifically, it fails to account for fat mass distribution in different body sites and neglects the variance in human height \cite{nuttall2015body}. On top of these issues, BMI suffers from age, gender, and ethnic biases \cite{deurenberg1998body}. These pitfalls have motivated research towards better body composition indicators that utilize accessible patient features, such as waist circumference and waist-to-height ratio \cite{lee2008indices, huxley2010body}, which can be measured in low-resource settings. However, most research often relies on black-box modeling that does not show the explicit relationship between predictors and outcomes of interest \cite{ferenci2018predicting}. Our experiments show that we can more accurately predict total body fat percentage through SR approaches, while maintaining explicit, interpretable model representation. 

\section{Symbolic Regression Overview}
\label{sr_overview}

The task of SR is to find an optimal mathematical function to estimate a given target variable. Specifically, SR seeks to learn a mapping $\hat{y}(\vec{x}) = f^{*}(\vec{x}): \mathbb{R} \mapsto \mathbb{R}^d$, assuming that a ground-truth solution ${y}(\vec{x}) = f(\vec{x})$ exists, where $\vec{x} \in \mathbb{R}^d$ are the available features and $y$ is the target. The optimal function $f^{*}$ can be composed of any subset of features in $\vec{x}$ and any set of mathematical operators that transform or fuse the features together. Finding an optimal SR model is not without its challenges. SR is fundamentally an NP-hard problem \cite{virgolin2022NPHARD}, in that the search space is infinite; the optimal symbolic expression can theoretically reach any length. Meanwhile, the options for mathematical operations can extend in theory to customized design. Despite these challenges, performing symbolic regression has been achieved through a variety of approaches, outlined in the following subsections. 

\subsection{Genetic Programming Approaches}

\begin{figure*}[!htp]
    \centering
    \includegraphics[width=\textwidth]{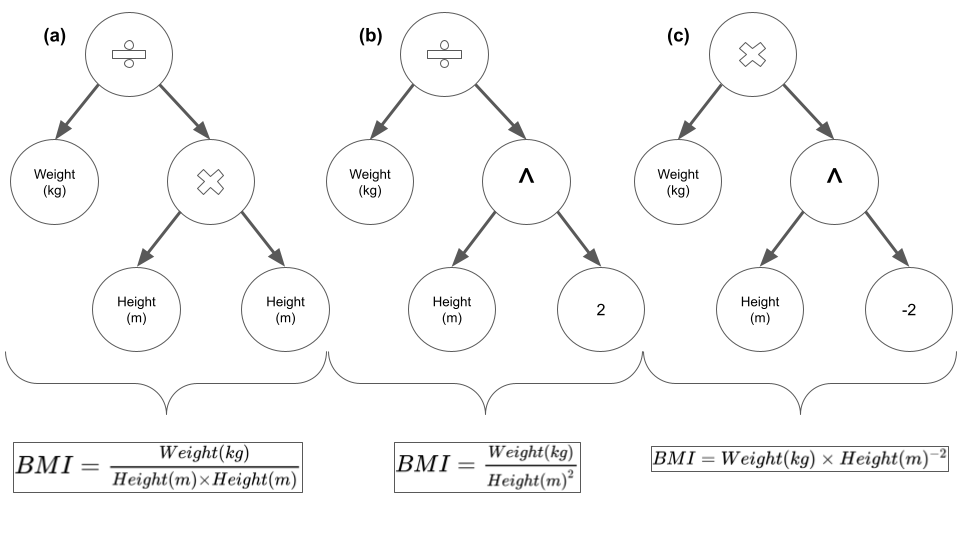}
    \caption{BMI Formula Tree Syntax Example}
    \label{fig:bmi_examnple}
\end{figure*}

Most commonly, genetic programming (GP) \cite{koza1992genetic} is leveraged to learn an SR model by randomly generating a set of candidate expressions and then gradually improving the candidates through series of mutation, crossover, reproduction, and selection operations, until the best candidate model satisfactorily fits the designated target variable. In these cases, the SR model is represented as a syntax tree, in which internal nodes represent mathematical operators, while terminal nodes represent features or constants, which neatly collapses into a single mathematical expression via recursive tree traversal. Figure \ref{fig:bmi_examnple} shows a few examples of syntax trees that could be used to represent the well-known BMI formula (Equation \ref{eqn:oldBMIeqn}), along with their corresponding expressions, demonstrating the variety in equivalent symbolic representation. 

In traditional Koza-style genetic programming approaches \cite{stephens_2016}, the initial population of candidate syntax trees are naively generated by randomly combining features, mathematical operators, and constants, such that each tree resembles a coherent mathematical expression. Then, a subset of trees are selected to proceed to the next generation via fitness competition, defined by each candidate's prediction error or correlation with respect to the target variable. Each selected tree undergoes genetic operations according to certain predefined probabilities, which can be empirically tuned. Mutation refers to the random change in either a single tree node (e.g. a terminal node is replaced by a random feature or constant, or an internal node is replaced by a random mathematical operator) or an entire subtree (e.g. replacing a subtree with a terminal node of the same subtree). Crossover refers to the random replacement of a subtree with a subtree from a donor. Reproduction refers to the cloning of a tree, resulting in a duplicate. After all selected trees have undergone genetic operations, they proceed to the next generation and fitness is evaluated again. The process repeats until some stopping criteria is met, either through a predictive performance evaluation or a maximum generation threshold.

The benefits of genetic approaches include limited \textit{a priori} assumptions about model structure, in addition to natural feature selection and engineering. However, the traditional approaches, in particular, suffer from high computational demands and difficulties in finding an appropriate trade-off between accuracy and complexity. As a result, other genetic strategies, such as Pareto-front exploitation methods \cite{smits2005pareto, deb2000fast, bleuler2001multiobjective}, seek to optimize multiple competing objectives, specifically complexity and performance, via hierarchical dominance criteria, considering both dimensions as equally important. This approach has been further improved by accounting for environmental and mating selection \cite{zitzler2001spea2} to better guide the search toward the Pareto-optimal front while maintaining storage space. Meanwhile, other Pareto optimization methods account for the age of an expression \cite{schmidt2011age, hornby2006alps, vladislavleva2008order} to avoid premature convergence. 

Other GP approaches include semantics-based methods that focus on the output values of candidate expressions, rather than the tree syntax itself. These methods prioritize better parent selection \cite{la2016epsilon, la2018learning} or integrate backpropagation into the learning process, especially for constant optimization \cite{topchy2001faster, kommenda2013effects, virgolin2019linear, burlacu2020operon}, which is not well-handled in traditional GP that depends on random mutations for model alterations. While GP techniques are generally lauded for learning model structure directly from the data, some GP approaches make assumptions about model formulation prior to training; these methods are closely related to generalized linear models that consider an assembly of potentially nonlinear features derived from the data \cite{MRGP, searson2015gptips, EFS, de2021interaction, huynh2018genetic}. Other novel GP approaches perform memetic variation by exploiting linkage information (strong inter-dependencies), thereby combining GP with information theory \cite{GOMEA, GPGOMEA}, resulting in performance that has been benchmarked at the Pareto-optimal front \cite{la2021contemporary}. More information about GP can be found in \textit{A Field Guide for Genetic Programming} \cite{poli08:fieldguide}.

\subsection{Other Approaches}
Some methods external to GP take inspiration from the traditional approaches but aim to develop SR as an efficient technology by generating simple models through the combination of randomly generated basis functions as features whose coefficients are tuned via pathwise regularized learning \cite{mcconaghy2011ffx} or by only preserving elite bases for a generalized linear model \cite{chen2017elite}. Other approaches take inspiration from tree structure representation alone and use Markov Chain Monte Carlo (MCMC) Bayesian probabilistic methods \cite{jin2019bayesian} to assign posterior distributions to said structures; the final model is linear combination of small trees. Neural network approaches have also been used to extract expressions with feed-forward architectures \cite{martius2016extrapolation,  sahoo2018learning, kim2020integration}, while others generate conditional production rules for tree grammars 
with the help of Monte Carlo Tree Search (MCTS) for guidance in the search space \cite{li2019neural}. Deep reinforcement learning has also been implemented using a large recurrent neural network to search the space of small models via a risk-seeking policy gradient \cite{petersen2019deep}. Although GP approaches have existed as the dominant SR technique for many years, non-genetic metalearning strategies have proven to be competitive on a variety of problems and hold promise for future SR model development. 

QLattice \cite{brolos2021approach} is an SR approach inspired by Richard Feynman's path integral formulation. Models are learned by tuning probability distributions associated with graphs that resemble mathematical functions (similar to syntax trees from genetic programming approaches). The QLattice simulates paths from inputs to outputs and randomly samples ``interactions,'' or transformations of inputs, reinforcing paths that provide high predictive utility. In this paper, we use the QLattice approach implemented by the Feyn library \cite{Feyn} for our experiments, due to its proven competitive performance on both synthetic and black-box problems \cite{SRbench}, as well as simplicity of implementation. 

\section{Experiments}

Motivated by the need for better body composition heuristics that utilize accessible patient features, in this case study, we generate SR models to improve estimation of total body fat percentage over the standard BMI heuristic and other baseline methods. The following subsections detail the experimental settings.  

\subsection{Data Collection and Preprocessing}
Our analysis utilizes a subset of the 2017-2018 NHANES database \cite{Nhanes}. For the purposes of this study, subjects were only included in the analysis if they were at least 18 years old, not pregnant at the time of examination, and possessed non-missing values within the following datasets: Demographic Variables and Sample Weights, Body Measures, and Dual-Energy X-ray Absorptiometry - Whole Body. The datasets were joined using the anonymous sequence numbers (``SEQN'') that represent the individuals sampled from the United States population. Features used for analysis are included in Table \ref{tab:descriptive_stats}, keeping the naming scheme consistent with the NHANES database. The designated target variable is total body fat \%, obtained via dual-energy x-ray absorptiometry, labeled ``DXDTOPF'' in the NHANES database. In total, $2403$ unique individuals were included in the final dataset used for analysis, $1158$ $(48.2\%)$ of which were male and $1245$ $(51.8\%)$ were female. The data was randomly partitioned into training and testing sets using $80:20$ split. 

{
\begin{table*}[!htp]
\resizebox{\textwidth}{!}{
\begin{tabular}{lllllllll}
\begin{tabular}[c]{@{}l@{}}NHANES \\ Variable\end{tabular} & \begin{tabular}[c]{@{}l@{}}Variable \\ Description\end{tabular} & Mean                   & Std.                   & Min.                   & 25\%tile               & 50\%tile               & 75\%tile               & Max.                   \\ \hline
SEQN                                                             & Anonymous ID Number                                             & \multicolumn{1}{c}{--} & \multicolumn{1}{c}{--} & \multicolumn{1}{c}{--} & \multicolumn{1}{c}{--} & \multicolumn{1}{c}{--} & \multicolumn{1}{c}{--} & \multicolumn{1}{c}{--} \\
RIAGENDR                                                         & Gender (1='M', 0='F')                                                          & \multicolumn{1}{c}{--} & \multicolumn{1}{c}{--} & \multicolumn{1}{c}{--} & \multicolumn{1}{c}{--} & \multicolumn{1}{c}{--} & \multicolumn{1}{c}{--} & \multicolumn{1}{c}{--} \\
RIDAGEYR                                                         & Age (years)                                                     & 38.1                   & 12.6                   & 18.0                   & 27.0                   & 38.0                   & 49.0                   & 59.0                   \\
BMXWT                                                            & Weight (kg)                                                     & 79.7                   & 20.4                   & 36.2                   & 64.9                   & 76.9                   & 91.9                   & 176.5                  \\
BMXHT                                                            & Height (cm)                                                     & 166.6                  & 9.3                    & 138.3                  & 159.4                  & 166.5                  & 173.8                  & 190.2                  \\
BMXLEG                                                           & Upper Leg Length (cm)                                           & 39.5                   & 3.6                    & 26.0                   & 37.0                   & 39.5                   & 42.0                   & 50.0                   \\
BMXARML                                                          & Upper Arm Length (cm)                                           & 37.0                   & 2.7                    & 29.6                   & 35.0                   & 37.0                   & 39.0                   & 45.5                   \\
BMXARMC                                                          & Arm Circumference (cm)                                          & 33.1                   & 5.1                    & 20.7                   & 29.4                   & 32.9                   & 36.4                   & 52.7                   \\
BMXWAIST                                                         & Waist Circumference (cm)                                        & 96.0                   & 16.3                   & 56.4                   & 83.8                   & 94.7                   & 106.4                  & 154.9                  \\
BMXHIP                                                           & Hip Circumference (cm)                                          & 104.6                  & 12.8                   & 77.8                   & 95.5                   & 102.7                  & 111.6                  & 168.5                  \\
\textbf{DXDTOPF}                                                 & \textbf{Total Body Fat \%}                                      & \textbf{33.1}          & \textbf{8.6}           & \textbf{12.1}          & \textbf{27.1}          & \textbf{32.9}          & \textbf{40.2}          & \textbf{56.1}         
\end{tabular}
}
\centering
\caption{Variable names and descriptions included in the analysis with descriptive statistics. The target variable is shown in bold.}
\label{tab:descriptive_stats}
\end{table*}
}

\subsection{Modeling Methods}
In order to evaluate the efficacy of the SR modeling approach, we first established four baseline linear regression models, learned using ordinary least squares via the \textit{statsmodels} library \cite{seabold2010statsmodels}, to estimate total body fat percentage:

\begin{itemize}
    \item Baseline 1 takes height and weight as separate features.
    \item Baseline 2 takes as input the standard BMI formula (Equation \ref{eqn:oldBMIeqn}) as the only input feature.
    \item Baseline 3 takes all available body measurement features as inputs, specifically weight, height, upper leg length, upper arm length, arm circumference, waist circumference, and hip circumference.
    \item Baseline 4 includes all previously mentioned body measurement features, in addition to gender and age.
\end{itemize}

\begin{figure}[hb]
    \centering
    \includegraphics[width=0.45\textwidth]{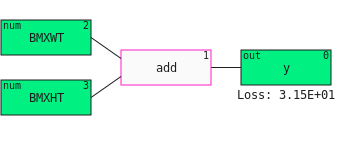}
    \caption{Baseline 1 Model, QLattice graph representation.}
    \label{fig:Baseline1}
\end{figure}

To compare against the four baseline models, we generated four competing SR models using the same input features as the baselines. The SR models were learned using the QLattice approach implemented by the Feyn library \cite{Feyn}, using the default parameter settings of Feyn's ``auto\_run'' function, with the exceptions of the ``max\_complexity'' parameter (the number of interactions, or edges in the graph representation of the model) and the number of training epochs; each SR model was allowed a maximum complexity as its corresponding linear baseline model. For example, to compete against baseline 1, whose graph representation is shown in Figure \ref{fig:Baseline1}, we generate a SR model using only height and weight features as input and a maximum complexity of 3, corresponding to the number of edges in the graph representation. A similar process was followed for the remaining baselines. All SR models were allowed to train for 100 epochs (as opposed to the default 10 epochs) for thorough learning and were evaluated using $R^2$ score on the training and testing sets.

\subsection{Results}
The baseline linear regression models and SR models are shown in Table \ref{tab:performance} with their respective performance on training and testing sets. The QLattice was able to learn SR models with equal or better predictive performance than the established baselines with respect to $R^2$ score on the test set. When SR model complexity was restricted to a maximum of 3 edges and only 2 input features, height and weight, the QLattice determined that the optimal model (SR Model 1) was almost equivalent to Baseline 1, a simple linear regression model. Increasing the maximum complexity to 4 edges (SR Model 2) provided minimal improvement, implying that height and weight features alone may not provide sufficient predictive power for estimating total body fat percentage. Interestingly, the traditional BMI formula produced the model with the worst predictive performance of all baselines and competing SR models, achieving an $R^2$ value of only 0.358 on the test set.

{
\newcommand{\zerodisplayskips}{%
  \setlength{\abovedisplayskip}{3pt}%
  \setlength{\belowdisplayskip}{3pt}%
  \setlength{\abovedisplayshortskip}{1pt}%
  \setlength{\belowdisplayshortskip}{1pt}}
\appto{\normalsize}{\zerodisplayskips}
\appto{\small}{\zerodisplayskips}
\appto{\footnotesize}{\zerodisplayskips}


\begin{table*}[!ht]
\resizebox{\textwidth}{!}{
    \begin{tabular}{|l|l|p{42em}|c|c|c|c|} 
    \hline
    \textbf{\begin{tabular}[c]{@{}l@{}}Model \\ Name\end{tabular}} 
    & \textbf{\begin{tabular}[c]{@{}l@{}}Input \\ Variables\end{tabular}} 
    & \textbf{Model} 
    & \textbf{\begin{tabular}[c]{@{}l@{}}Maximum \\ Complexity\end{tabular}} 
    & \textbf{\begin{tabular}[c]{@{}l@{}}Actual \\ Complexity\end{tabular}}
    & \textbf{\begin{tabular}[c]{@{}c@{}}Train \\ $R^2$\end{tabular}} 
    & \textbf{\begin{tabular}[c]{@{}c@{}}Test \\ $R^2$\end{tabular}} \\ 
    \hline \hline
    Baseline 1 
        & \begin{tabular}[c]{@{}l@{}}BMXWT, \\ BMXHT\end{tabular} 
        & \parbox{42em}{
            \begin{gather*}
                \hat{y} = 0.264311(BMXWT) -0.696876(BMXHT) + 128.138627
            \end{gather*}
        }
        & -- 
        & 3 
        & 0.563
        & 0.590 \\ 
        \hline
    Baseline 2 
        & \begin{tabular}[c]{@{}l@{}}BMXWT, \\ BMXHT\end{tabular} 
        & \parbox{42em}{
            \begin{gather*}
                \hat{y} = 0.733466 \left(\frac{BMXWT}{(0.01(BMXHT))^2} \right) + 12.084282
            \end{gather*}
        }
        & --
        & 4 
        & 0.329 
        & 0.358 \\ 
        \hline
    Baseline 3 
        & \begin{tabular}[c]{@{}l@{}}BMXWT, \\ BMXHT, \\ BMXLEG, \\ BMXARML, \\ BMXARMC, \\ BMXWAIST, \\ BMXHIP\end{tabular} 
        & \parbox{42em}{
            \begin{gather*}
                \hat{y} = -0.312160(BMXWT) -0.237978(BMXHT) -0.109314(BMXLEG) ...\\ 
                \hspace{4em} + 0.003146(BMXARML) -0.123752(BMXARMC) ...\\ 
                \hspace{4em} + 0.248836(BMXWAIST) + 0.635005(BMXHIP) ...\\ 
                \hspace{4em} + 15.689953 
            \end{gather*}
        }
        & --
        & 13 
        & 0.737 
        & 0.776 \\ 
        \hline
    Baseline 4 
        & \begin{tabular}[c]{@{}l@{}} RIAGENDR, \\ RIDAGEYR, \\ BMXWT, \\ BMXHT, \\ BMXLEG, \\ BMXARML, \\ BMXARMC, \\ BMXWAIST, \\ BMXHIP\end{tabular} 
        & \parbox{42em}{
            \begin{gather*}
                \hat{y} = -0.169817(BMXWT)  -0.103678(BMXHT) + 0.075362(BMXLEG)  ...\\ 
                \hspace{4em} - 0.046780 (BMXARML) +  0.069824(BMXARMC)  ...\\ 
                \hspace{4em} + 0.318612(BMXWAIST) + 0.249766(BMXHIP) ...\\ 
                \hspace{4em} + 8.675876(RIAGENDR) + 0.007782(RIDAGEYR) ...\\ 
                \hspace{4em} -1.087157
            \end{gather*}
            \vspace{0pt}
        }
        & --
        & 17 
        & 0.820 
        & 0.843 \\ 
        \hline \hline
    SR Model 1 
        & \begin{tabular}[c]{@{}l@{}}BMXWT, \\ BMXHT\end{tabular} 
        & \parbox{42em}{
            \begin{gather*}
                \hat{y} = 0.264355(BMXWT) - 0.697227(BMXHT) + 128.221
            \end{gather*}
            }
        & 3
        & 3 
        & 0.563 
        & 0.590 \\ 
        \hline
    SR Model 2 
        & \begin{tabular}[c]{@{}l@{}}BMXWT, \\ BMXHT\end{tabular} 
        & \parbox{42em}{
            \begin{gather*}
                \hat{y} = - 0.711635(BMXHT) + 20.7829 \sqrt{0.0338353(BMXWT) - 1} + 125.119
            \end{gather*}
        }
        & 4
        & 4 
        & 0.569
        & 0.603 \\ 
        \hline
    SR Model 3 
        & \begin{tabular}[c]{@{}l@{}}BMXWT, \\ BMXHT, \\ BMXLEG, \\ BMXARML, \\ BMXARMC, \\ BMXWAIST, \\ BMXHIP\end{tabular} 
        & \parbox{42em}{
            \begin{gather*}
                \hat{y} = 533.592 e^{- 0.0382652 (BMXWAIST)} ...\\
                \hspace{4em} \times \Biggl(1.34147 - 0.0076698 (BMXHT) - 0.0124393 (BMXLEG) ...\\ 
                \hspace{8em} + \biggl(0.0731049 (BMXWAIST) - 3.98442\biggr) ...\\ 
                \hspace{8em} \times \biggl(0.0212026 (BMXHIP) ...
                - 0.012399 (BMXWT) - 1.35696\biggr) \Biggr) \\ 
                \hspace{2em} + 44.9658.
            \end{gather*}
        }
        & 13
        & 12 
        & 0.791
        & 0.833 \\ 
        \hline   
    SR Model 4 
        & \begin{tabular}[c]{@{}l@{}} RIAGENDR, \\ RIDAGEYR, \\ BMXWT, \\ BMXHT, \\ BMXLEG, \\ BMXARML, \\ BMXARMC, \\ BMXWAIST, \\ BMXHIP\end{tabular} 
        & \parbox{42em}{
            \begin{gather*}
                \hat{y} = 43.3409 - 10.0751 \Biggl(2.23561 - 0.0130747(BMXWAIST)\Biggr) ...\\
                \hspace{4em} \times \Biggl(0.0174372(BMXWAIST) 
                + \biggl(0.0306655(BMXHIP) - 5.21981\biggr) ...\\ 
                \hspace{8em} \times \biggl(0.019191(BMXWAIST) - 0.0146988(BMXWT) - 2.03625\biggr) ... \\ 
                \hspace{8em} + \biggl(RIAGENDR_{cat} - 0.164541\biggr) ... \\
                \hspace{8em} \times \biggl(1.19881 \Bigl(2.62601 - 0.0694562(BMXARMC)\Bigr) ...\\
                \hspace{12em} \times \Bigl(4.57897 - 0.0338513(BMXHT)\Bigr) + 2.14167\biggr) 
               - 3.16342\Biggr),
            \end{gather*}
            \[
                \textit{where} \hspace{2em} RIAGENDR_{cat} = 
                \begin{cases} 
                    -0.2514210227924248,  & \textit{Gender = Male} \\
                    0.24145479395502106,  & \textit{Gender = Female} 
                \end{cases}
             \] 
        }
        & 17
        & 17 
        & \textbf{0.856} 
        & \textbf{0.879} \\ 
        \hline
    \end{tabular}
}
\caption{Baseline linear regression models and SR models with their respective performance on training and testing sets. The best model $R^2$ score is shown in bold.}
\label{tab:performance}
\end{table*}
}
        
\begin{figure*}[!ht]
    \centering
    \includegraphics[width=\textwidth]{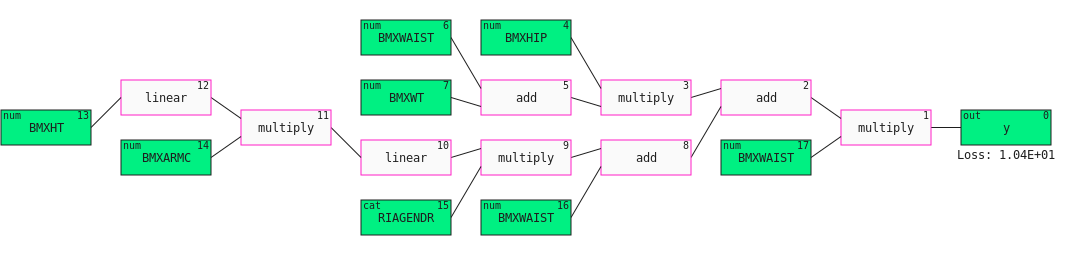}
    \caption{SR Model 4, QLattice graph representation.}
    \label{fig:SRmodel4}
\end{figure*}

Baseline performance improved significantly by providing all available body measurements as features (Baseline 3), and it improved even further with the additions of gender and age (Baseline 4). The same was true for SR Models 3 and 4, which both outperformed their respective baselines. The best SR Model (SR Model 4), whose graph representation is shown in Figure \ref{fig:SRmodel4}, was allowed a maximum complexity of 17 edges, achieving an $R^2$ value of $0.879$ on the test set. Importantly, all resulting symbolic expressions are explicit and interpretable. Although predictive performance (or correctness) is not a component of interpretability, prior to further discussion, we have demonstrated that the SR approach can provide a more robust formula than the rudimentary linear models to estimate body fat percentage, while maintaining the same level of complexity. We will continue to analyze SR Model 4 in the following sections.

\section{SR Model Interpretability}

A high level of interpretability supports the evaluation of successful information fusion, in that interpretable models tend to show explicit relationships and interactions between features and outputs, allowing for in-depth analysis of the knowledge gained through ML and how information concretely synthesizes to reveal important patterns. Human comprehensibility is not only important for fidelity in high-risk situations with unexpected predictions \cite{Lavrac1999SelectedTF} but also for its downstream effects, such as the formulation of new theories or hypotheses in scientific domains \cite{Freitas2014}. In this overview, we demonstrate the interpretability of SR modeling in terms of previously established evaluation methods. 

Importantly, we differentiate between the concepts of \textit{explainability} and \textit{interpretability}. Explainability most often relies on post-hoc analysis of ``black box'' ML models by fitting secondary models to explain the primary model \cite{explainableSR}, which are often not fully representative of the original \cite{rudin2019stop}; whereas interpretability, while acknowledging that this topic is still widely disputed in the literature \cite{LiptonMyth, bibal2016interpretability}, more readily provides insight into the learned relationships \cite{Murdoch22071} because the model generates explanations in the process of decision making or while being trained \cite{causal}. The challenge accompanying interpretability is the potential trade-off between human understanding and model performance, or descriptive accuracy vs. predictive accuracy desiderata \cite{Murdoch22071}. However, this trade-off is not always observed when meaningful features have been constructed \cite{rudin2019stop}. Some studies have developed taxonomies \cite{doshi2017towards} for evaluating model interpretability through application-grounded, human-grounded, and functionally-grounded analyses. While some argue that human expert evaluation via surveys is the ideal measure of interpretability, such research is limited by expert time and accessibility, along with the difficulty of meaningful survey construction  \cite{Rping2006LearningIM}. For the purposes of evaluating the interpretability of SR modeling, and therefore its support for the assessment of information fusion, we consider functionally-grounded, or heuristic,  measures of interpretability \cite{lage2019evaluation}, which are best demonstrated through model representation \cite{bibal2016interpretability}.

\subsection{Size, Sparsity, \& Simplicity}
The size of a model is a common heuristic used to evaluate its complexity, operating under the assumption that smaller models lead to higher levels of human comprehensibility \cite{bibal2016interpretability, Rping2006LearningIM}; Proponents of this argument often cite psychological studies that demonstrate humans can handle at most $7 \pm 2$ cognitive entities at once \cite{Miller1956TheMN}. However, the size of a model is a highly syntactical issue that fails to capture semantics \cite{Freitas2014}; one way to reduce a user's interpretation workload may involve focusing on the subset of most ``interesting'' (novel or surprising) patterns \cite{freitas2006we} that the model has found. On the other hand, models that appear to oversimplify problems to a higher degree than expected might lead to lower fidelity or acceptance, especially in fields such as medicine \cite{Lavrac1999SelectedTF}. Given this context, the principle of Occam's Razor \cite{Domingos1998OccamsTR}, most often embodied by the Minimum Description Length (MDL) principle \cite{Grunwald2007}, which prioritizes data compression in machine learning applications, states that a simpler model is often preferred when there is little trade-off in performance or generalizability. This concept is closely related to sparsity, which aims to limit the number of non-zero parameters, or signals, that factor into predictions, yielding higher descriptive accuracy \cite{Murdoch22071}. The trade-off between complexity and performance is the crucial component to consider when searching for an ideal symbolic expression to represent phenomena of interest. 

A common goal among SR approaches is to produce mathematical expressions that are as concise as possible, while maintaining high performance. In GP-based SR, this goal is often achieved by including a parsimony hyperparameter, aimed at punishing large syntactical tree depths and favoring smaller candidate model representations during the selection procedure, provided that they do not hinder performance to a great extent. To address this trade-off, some GP methods exploit Pareto dominance, which is the multi-objective optimization problem that places value in expressions that are ``non-dominated'' in at least one of the two dimensions (complexity and performance) in order to qualify for selection in next generation \cite{smits2005pareto, deb2000fast, bleuler2001multiobjective}.

The QLattice, in particular, includes a maximum complexity hyperparameter, which is designed to eliminate large models from the search space, as defined by the number of edges in a model's graph representation. In our experiments, we demonstrated that the QLattice was able to find symbolic expressions with equal or lower complexity than the linear baseline models without sacrificing performance on the test set, justifying the choice of the SR models over the baselines. 

\begin{figure*}[!ht]
    \centering
    \includegraphics[width=\textwidth]{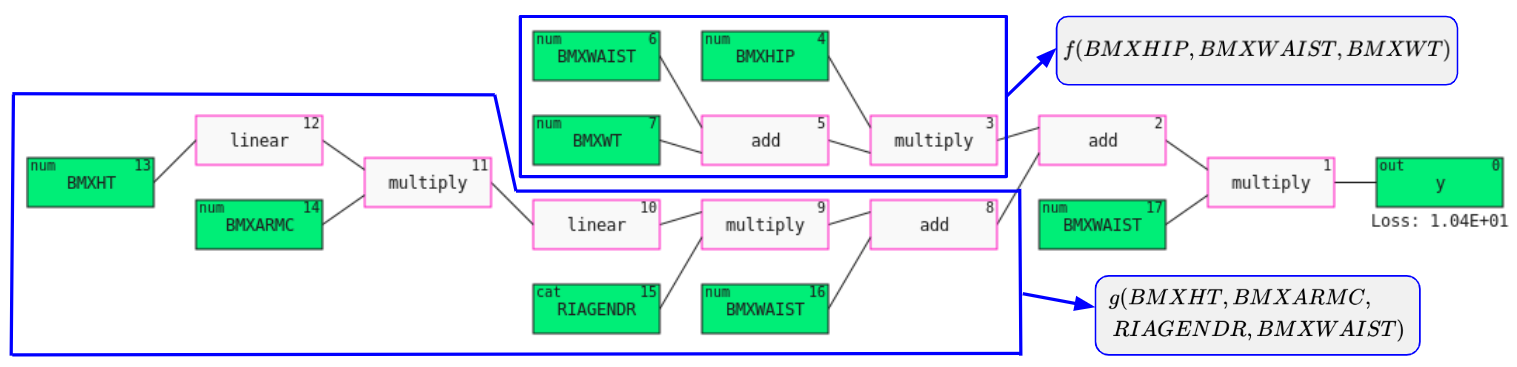}
    \caption{SR Model 4 graph representation with subgraphs boxed in blue, indicating areas for sub-analysis.}
    \label{fig:subtrees}
\end{figure*}

\subsection{Modularity \& Feature Engineering}
 Modularity refers to the ability to break up a model into meaningful portions \cite{Murdoch22071}, closely related to the concept of cognitive chunks \cite{doshi2017towards}, which are individual elements of information that can be independently analyzed. Since SR expressions are often represented as graph structures, it's possible to analyze subgraph expressions individually to glean insight into individual pieces of the model. For instance, within SR Model 4, we might consider compartmentalizing and separately analyzing certain subexpressions that factor into the final estimation of total body fat percentage. Figure \ref{fig:subtrees} shows how one might consider splitting the expression into modules, allowing us to rewrite the model more concisely:

{\small 
\begin{align*}
    \hat{y} =- 10.0751 \bigl(2.23561 - 0.0130747 (BMXWAIST)\bigr)  \bigl(f + g\bigr) + 43.3409,
\end{align*}
}

\noindent where
{
\begin{align*}
    \\
     f = \bigl(0.019191 (BMXWAIST) - 0.0146988 (BMXWT) - 2.03625\bigr)... \\ \times \bigl(0.0306655 (BMXHIP) - 5.21981\bigr), \\
\end{align*}
}
\noindent and 
{
\begin{align*}
    \\
     g = 0.0174372 (BMXWAIST) + \biggl(RIAGENDR_{cat} - 0.164541\biggr) ...\\
     \times \biggl(1.19881 \Bigl(2.62601 - 0.0694562(BMXARMC)\Bigr)... \\
     \times \Bigl(4.57897 - 0.0338513(BMXHT)\Bigr) + 2.14167\biggr)... \\
     - 3.16342.
\end{align*}
}

\begin{figure}[!hbp]
    \centering
    \includegraphics[width=0.5\textwidth]{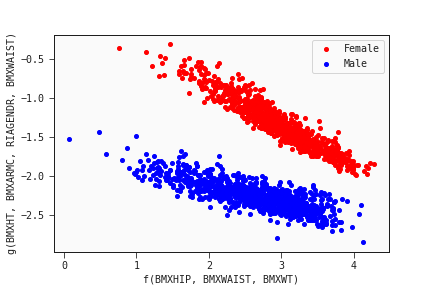}
    \caption{Visualization of model sub-components, $f$ vs. $g$ on training set.}
    \label{fig:fxgx}
\end{figure}

\noindent We can now analyze the subcomponents $f$ and $g$ through visualization, as shown in Figure \ref{fig:fxgx}, which shows a clear inverse linear relationship, well-separated by gender. 

 Certain approaches for SR intentionally formulate the learned expressions as a generalized additive model, or linear combinations of potentially non-linear trees \cite{MRGP, searson2015gptips, EFS, de2021interaction, huynh2018genetic}, which can easily be broken apart at junctures of addition. The result of this splitting procedure is the formation of meta-variables, which can be independently analyzed or used to simplify future modeling within similar problems \cite{smits2005pareto}. For SR Model 4, we may decide to mathematically expand the model to identify each term, which allows us to observe explicit interactions between features:

{ 
\begin{align*}
     \hat{y} = 
     0.000371294 (BMXARMC) (BMXHT) (BMXWAIST)... \\
     \hspace{2em} \times (RIAGENDR_{cat}) \\
     \hspace{2em} - 6.10931 \cdot 10^{-5} (BMXARMC) (BMXHT) (BMXWAIST) \\
     \hspace{2em} - 0.0634866 (BMXARMC) (BMXHT) (RIAGENDR_{cat}) \\
     \hspace{2em} + 0.0104462 (BMXARMC) (BMXHT) \\
     \hspace{2em} - 0.0502239 (BMXARMC) (BMXWAIST) (RIAGENDR_{cat}) \\
     \hspace{2em} + 0.00826389 (BMXARMC) (BMXWAIST) \\
     \hspace{2em} + 8.58766 (BMXARMC) (RIAGENDR_{cat}) \\
     \hspace{2em} - 1.41302 (BMXARMC) \\
     \hspace{2em} + 7.75227 \cdot 10^{-5} (BMXHIP) (BMXWAIST^{2}) \\
     \hspace{2em} - 5.93763 \cdot 10^{-5} (BMXHIP) (BMXWAIST) (BMXWT) \\
     \hspace{2em} - 0.0214809 (BMXHIP) (BMXWAIST) \\
     \hspace{2em} + 0.0101526 (BMXHIP) (BMXWT) \\
     \hspace{2em} + 1.40646 (BMXHIP) \\
     \hspace{2em} - 0.0140379 (BMXHT) (BMXWAIST) (RIAGENDR_{cat}) \\
     \hspace{2em} + 0.00230982 (BMXHT) (BMXWAIST) \\
     \hspace{2em} + 2.40031 (BMXHT) (RIAGENDR_{cat}) \\
     \hspace{2em} - 0.39495 (BMXHT)\\
     \hspace{2em} - 0.0108987 (BMXWAIST^{2}) \\
     \hspace{2em} + 0.0101069 (BMXWAIST) (BMXWT) \\ 
     \hspace{2em} + 2.18099 (BMXWAIST) (RIAGENDR_{cat}) \\
     \hspace{2em} + 2.4881 (BMXWAIST) \\
     \hspace{2em} - 1.72815 (BMXWT) \\ 
     \hspace{2em} - 372.922 (RIAGENDR_{cat}) \\ 
     \hspace{2em} - 63.4491.
\end{align*}
 }
 
 Regarding modularity, there is future potential within SR to express the model in terms of some global function plus local functions \cite{Hand2002PatternDA, Rping2006LearningIM}, in which the global expression may find obvious or simple patterns in the data, while the local pieces capture more interesting components specific to to certain subsets or cases within the feature space. For example, SR Model 4 may serve as a global approximation of body fat percentage; however, it does not account for the anthropometric differences between ethnic groups. For this problem it may be reasonable to search for additional expressions to capture patterns on subsets of the data, which may resemble
 
\begin{equation*}
    f(x) + \begin{cases} 
                g(x) + \epsilon,  & \textit{Ethnic Group = White or Caucasian} \\
                h(x) + \epsilon,  & \textit{Ethnic Group = Hispanic or Latino} \\
                \hspace{2em}. \\
                \hspace{2em}. \\
                \hspace{2em}. \\
            \end{cases}
\end{equation*}
 
 \noindent where $f(x)$ is our global function that captures broad phenomena, while $g(x)$ and $h(x)$ are local functions that capture case-based or subset-based nuances.

\subsection{Simulatability, Counterfactuals, \& Function Analysis}

Simulatability refers to the ability of the human user to directly mimic the model's decision-making process from input to output \cite{Murdoch22071}. Since the end product of the SR method is essentially a formula, simulatability boils down to the simple process of plugging values into the expression, provided that the user has access to all necessary information. The more interesting interpretability component associated with simulatability is related to counterfactual analysis \cite{hume2000enquiry, lewis1986causation}, which aims to study the state of affairs resulting from an event not occurring and often accompanies the study of causal inference \cite{Miller2019ExplanationIA}. In plain terms, counterfactuals deal with questions of the ``What if?'' variety (e.g. ``how will body fat percentage estimation change if the individual was 10 centimeters shorter, while keeping all other variables the same?''). Function analysis is the obvious natural extension of this concept, and it is the main reason that SR's degree of interpretability is high; the post-hoc analysis of SR does not depend on secondary model fitting to explain predictions. Additionally, it does not require case-based demonstration for in-depth understanding like decision trees \cite{breiman2017classification} or rule ensembles \cite{friedman2008predictive} require. Instead, data scientists or domain experts can use standard mathematical analysis techniques to understand how the model explicitly works. Using only the model representation itself, we may find intercepts or roots, analyze asymptotic behavior, and compute derivatives and integrals. The principles of calculus allow us to determine critical points, increasing and decreasing intervals, extrema, concavity, and inflection points. 

To illustrate this point, in SR Model 4, we can observe that waist circumference plays a crucial role in total body fat percentage estimation. In a practical use case for a fully grown adult, we may want to know how quickly total body fat percentage may change in response to a change in waist circumference for the average male vs. the average female. Using calculus, we can compute partial derivative of the function with respect to waist circumference:

{
\begin{flalign*}
    \frac{\partial \left[ \textit{SR Model 4}\right]}{\partial BMXWAIST}
    = 
    0.00229698 (BMXWAIST) + ... \\
    \hspace{4 em} + \Bigl(0.000588502 (BMXHIP) - 0.0827362\Bigr)... \\
    \hspace{12 em} \times \Bigl( 0.131729 (BMXWAIST) - 22.524\Bigr)... \\
    \hspace{5 em} + 0.131729 \Bigl(0.0306655 (BMXHIP) - 5.21981\Bigr)... \\ \times \Bigl(0.019191 (BMXWAIST) - 0.0146988 (BMXWT) - 2.03625\Bigr) \\
    \hspace{4 em} + 0.131729 \biggl(RIAGENDR_{cat} - 0.164541\biggr) \\
    \hspace{10 em} \times \biggl(1.19881 \Bigl(2.62601 - 0.0694562 (BMXARMC)\Bigr)... \\
    \hspace{10 em} \times \Bigl(4.57897 - 0.0338513 (BMXHT)\Bigr)+ 2.14167\biggr)... \\ 
    \hspace{20 em} - 0.416714.
\end{flalign*}
}

\noindent Then, using this partial derivative, we can plug in the average body measurements for adult males and females respectively and determine the rate at which estimated body fat percentage might change for varied waist measurements. For instance, the partial derivative evaluated for the average female adult who weighs $73.95$ kg, with a height of $160.46$ cm, arm circumference of $31.87$ cm, and hip circumference of $106.13$ cm and the average male adult who weighs $85.97$ kg, with a height of $173.19$ cm, arm circumference of $34.33$ cm, and hip circumference of $102.95$ cm would be:

{
\begin{align*}
    \frac{\partial \left[ \textit{SR Model 4}\right]}{\partial BMXWAIST} \approx
    \begin{cases} 
        - 0.0053(BMXWAIST) + 0.866,  
            & \textit{Sex = 'Female'} \\ \\
        - 0.0058 (BMXWAIST) + 0.882,  
            & \textit{Sex = 'Male'}
    \end{cases}
\end{align*}
}

\begin{figure}[!ht]
    \centering
    \includegraphics[width=0.9\textwidth]{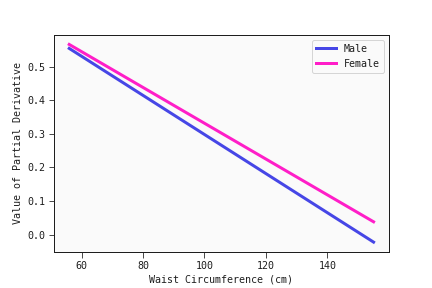}
    \caption{Partial derivative of SR Model 4 with respect to waist circumference (cm), evaluated for average male vs. average female adults}
    \label{fig:viz2}
\end{figure}

\noindent When we plot waist circumference (cm) vs. the value of the partial derivative, keeping all other variables constant at their average values for each gender, respectively, as shown in Figure \ref{fig:viz2}, we observe that the rate of change of total body fat percentage is linear and it is higher for the average female than for the average male for all possible values of waist circumference. This is an interesting finding for the dynamics of body fat percentage between sexes.

\section{Discussion}

In this work, we have demonstrated that SR techniques fuse important information into mathematical expressions that capture potential non-linear relationships hidden within data, while maintaining a high level of intrinsic interpretability. As new approaches for SR continue to emerge, the main benefits of this task will continue to hold promise for future applications that rely on highly portable, interpretable models, especially in high-stakes fields that require in-depth model analysis. 

Evaluating model interpretability in a strict sense has been hindered due to the numerous differing terms attributed to interpretability (i.e. the lack of consensus in the scientific literature); the result is that interpretability is difficult to precisely measure \cite{bibal2016interpretability}, and that which cannot be strictly measured cannot be strictly evaluated in a quantifiable way. However, we argue that symbolic mathematical notation, which is based on a human-created grammar that has persisted across a variety of languages, cultures, and eras, has enabled us to convey highly complex concepts in a remarkably concise way, creating a truly universal subject \cite{venezia2016development}. The idea that mathematical notation simultaneously represents both a concept and process has been understood for many years and has allowed for high-level cognitive manipulations that have played a fundamental role in the expansion of our collective knowledge base \cite{tall1992mathematical}. The expansion of knowledge is dependent not only on successful information fusion but the understanding of said fusion. Therefore, the pursuit to improve SR techniques, which produce models embodied by this mathematical representation, is a valuable one.


\section*{Acknowledgments}
\label{ack}
Funding: The research was funded by the University of Notre Dame's Lucy Family Institute for Data and Society. 

\bibliographystyle{elsarticle-num} 
\bibliography{main}





\end{document}